\documentclass{article}

\usepackage[preprint]{neurips_2026}

\usepackage[utf8]{inputenc}
\usepackage[T1]{fontenc}
\usepackage{hyperref}
\usepackage{url}
\usepackage{booktabs}
\usepackage{amsfonts}
\usepackage{amsmath}
\usepackage{nicefrac}
\usepackage{microtype}
\usepackage{xcolor}
\usepackage{graphicx}
\usepackage{multirow}

\title{Bits and Memories: Measuring Verbatim Extraction\\
Across LLM Quantization}

\author{%
  Akshay Sasi \\
  Independent Researcher \\
  \texttt{akshaysasi12.knr@gmail.com} \\
}

\begin{document}

\maketitle

\begin{abstract}
Language models are almost always quantized before they are deployed, and a
growing line of work asks whether quantization also lowers their privacy
risk. That work measures privacy almost entirely with membership inference.
We think this is the wrong thing to measure for the risk that most people
actually worry about, namely a model reproducing its training data word for
word, and we measure that directly. Using the Pythia models and the public
set of sequences each of them is known to have memorized, we track verbatim
extraction across five precision levels, from full precision down to four
bits, and across three model sizes, while measuring general capability
(perplexity) at every point. We find two things. Quantization is a selective
forgetter: verbatim memorization falls off faster than capability at every
precision and every model size we tried, and this holds under two unrelated
quantization algorithms and two evaluation corpora. But the selectivity is
not enough to make quantization a privacy defense, which cuts against the
optimistic reading of earlier membership-inference results. At the largest
model we study, four-bit quantization still reproduces most of the memorized
sequences while giving up only a few percent of capability, and the fraction
of memorized data that survives quantization \emph{grows} with model size. We conclude that
compression should not be treated as a way to remove memorized training
data, and that extraction, not membership inference, is the number
practitioners should be watching. All code, sampled evaluation data, and
per-configuration results are released.
\end{abstract}

\section{Introduction}

Two facts about large language models are now well established. Set side by
side, they should make anyone deploying a model a little nervous.

The first is that models memorize. Given the right prefix, a trained model
will often continue with a passage from its training set, character for
character. This has been demonstrated repeatedly, from the early extraction
attacks on GPT-2 \citep{carlini2021extracting} to systematic studies showing
that the amount of memorized text grows with model size, with how many times
a passage was duplicated in training, and with how much context the model is
given \citep{carlini2022quantifying}, and up to large-scale extraction from
production systems \citep{nasr2023scalable}. Memorized text is a liability:
it can be private, it can be copyrighted, and it can be pulled back out by
anyone who knows how to prompt for it.

The second fact is that almost every model that gets deployed is
\emph{quantized}. Storing weights at eight, four, or fewer bits instead of
sixteen or thirty-two is what makes it possible to run a capable model on a
single consumer GPU. The techniques are mature and widely used, from
eight-bit inference \citep{dettmers2022int8} to four-bit post-training
methods \citep{frantar2022gptq,dettmers2023qlora} and activation-aware
variants \citep{lin2023awq,xiao2023smoothquant}.

Put these together and a practical question falls out. When you quantize a
model, what happens to the data it memorized? If it mostly gets erased, then
quantization is a privacy bonus that essentially every deployment already
gets for free. If it survives, then the comfortable assumption that a
compressed model is a safer model is wrong, and someone should say so.

Recent work has started on this question, but almost always through the lens
of \emph{membership inference}: can an attacker tell whether a particular
example was in the training set? Several papers report that quantization
reduces membership-inference success, on code models
\citep{kumar2025quantcode}, on image classifiers \citep{anon2025bits}, and
across compression schemes more broadly \citep{li2026compleak}, and the
natural takeaway is that quantization helps privacy. We think this takeaway
is misleading, because membership inference is a weak stand-in for the threat
that generates lawsuits and headlines. Knowing that a document was in the
training data is very different from being able to \emph{print the document}.
Printing it back out is verbatim extraction, and as far as we can tell nobody
has measured how it responds to quantization, across bit widths and across
model scale, against ground-truth memorized data.

That is what this paper does.\footnote{Code: \url{https://github.com/AkshaySasi/bits-and-memories}. Sampled data and per-configuration results: \url{https://huggingface.co/datasets/AkshaySasi/bits-and-memories}.} We use the Pythia suite
\citep{biderman2023pythia}, the one family of open models whose training
corpus, the Pile \citep{gao2020pile}, and whose memorized sequences are both
public \citep{biderman2023emergent}, so that we never have to guess what a
model memorized. We use the official list. We prompt each model with the
first half of a known memorized sequence, let it greedily generate the
second half, and check for an exact match. We do this across five precision
levels and three model sizes, and, crucially, we measure general capability
(perplexity on held-out text) at every single point, so that we can tell
``the model forgot this passage'' apart from ``the model is simply broken.''

We have two findings, and they pull against each other, which is what makes
the result interesting to us.

\begin{enumerate}
  \item \textbf{Quantization forgets memorization faster than it forgets
    capability.} At every model size and every precision, verbatim
    extraction drops off faster than perplexity rises. We summarize this with
    a single number, a \emph{selectivity} ratio, and it is comfortably above
    one everywhere, reaching about seven at our largest model. This is
    robust: it appears under bitsandbytes' NF4 and under a plain,
    dependency-free round-to-nearest quantizer we wrote ourselves, and it
    appears whether we measure capability on WikiText or on in-distribution
    Pile text.
  \item \textbf{But quantization is not a privacy defense.} The selective
    forgetting is real, and it is still not enough. At our one-billion
    parameter model, four-bit quantization keeps roughly ninety-five percent
    of capability while still reproducing over seventy percent of memorized
    sequences. Worse for the optimistic story, the surviving fraction
    \emph{increases} with model size, from the smallest model to the largest.
    Since real deployed models are far larger than the ones we can run, the
    honest extrapolation is that compression leaves memorized data
    extractable.
\end{enumerate}

For anyone shipping a model, the takeaway is short: do not count on
quantization to scrub memorized training data, and watch extraction rather
than membership inference when you want to know your exposure. There is also
a less obvious point, which we come back to in the discussion. Memorized text
seems to sit in a more precision-fragile part of the weights than general
capability does, and larger models appear to shield their capability from
quantization noise faster than they shield individual memories.

\section{Related work}

\paragraph{Memorization and extraction.}
That language models reproduce training data was shown directly by
\citet{carlini2021extracting}, who extracted verbatim sequences, including
personal information, from GPT-2. \citet{carlini2022quantifying} turned this
into a quantitative science, establishing log-linear relationships between
memorization and model scale, duplication count, and context length.
\citet{biderman2023emergent} studied which sequences a model will memorize
and released, for the Pythia models, the ground-truth memorized sets we use
here. More recent work refines how extraction should even be defined and
measured, moving from strict greedy reproduction toward probabilistic notions
of discoverable extraction \citep{hayes2024measuring}, and studies
memorization at finer granularity \citep{zhou2023entity}. Extraction has also
been scaled up to production models \citep{nasr2023scalable}. We adopt the
strict greedy-reproduction definition because it is the most conservative and
the least ambiguous: an exact match is an exact match.

\paragraph{Quantization.}
Post-training quantization of language models is a large and fast-moving
area. Eight-bit inference with outlier handling \citep{dettmers2022int8}
made lossless eight-bit deployment routine. GPTQ \citep{frantar2022gptq}
brought accurate four-bit and lower post-training quantization using
second-order information; QLoRA \citep{dettmers2023qlora} introduced the NF4
data type we use as our main four-bit method; and activation-aware and
smoothing methods \citep{lin2023awq,xiao2023smoothquant} further reduced the
accuracy cost. Inference scaling laws show that four bits sits near the
accuracy-per-bit sweet spot \citep{dettmers2023kbit}, which is one reason it
is the regime deployments favor and the regime where our effect is
strongest. These papers, understandably, evaluate quantization on task
accuracy and perplexity. None of them ask what quantization does to
memorized data, which is the gap we fill.

\paragraph{Compression and privacy.}
The intersection is recent and, so far, dominated by membership inference.
\citet{kumar2025quantcode} study quantization and membership inference on
code models and report reduced attack effectiveness at lower precision;
\citet{anon2025bits} report large membership-inference reductions from
post-training quantization on image classifiers; \citet{li2026compleak} show
that compression can, under some access models, \emph{increase} leakage.
Theoretical and empirical treatments of membership-inference risk under
quantization are beginning to appear \citep{anon2025miqrisk}. On the
mitigation side, \citet{zhang2025pruning} show that pruning reduces
memorization, the sibling result to ours for a different form of compression.
Most relevant to our interpretation, \citet{zhang2024catastrophic} show that
quantizing a model that has been \emph{unlearned} can bring the supposedly
removed knowledge back, direct evidence that memorized information can be
stored in a quantization-robust way. Our work differs from all of these by
measuring verbatim extraction rather than membership inference, across bit
widths and model scale, with a capability control at every point. Membership
inference itself \citep{shokri2017membership} remains valuable, but it
answers a different, and for many purposes weaker, question. A broader survey
of language-model privacy risks is given by \citet{smith2023survey}.

\section{Method}
\label{sec:method}

\begin{figure}[t]
  \centering
  \includegraphics[width=\linewidth]{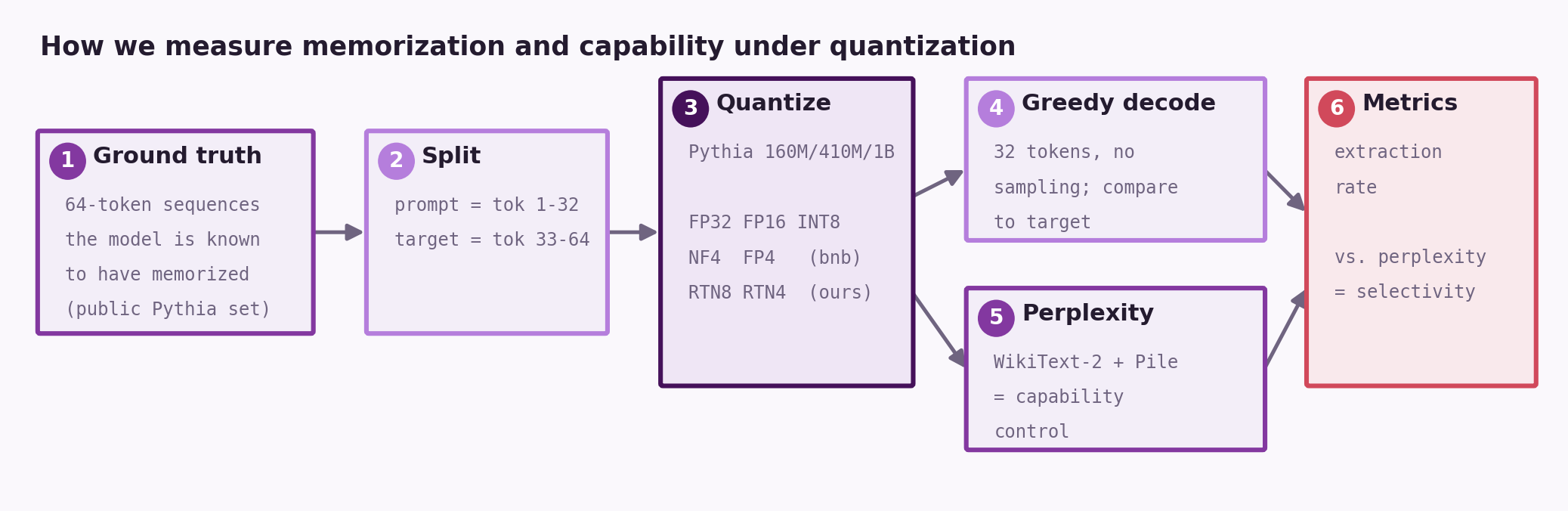}
  \caption{The measurement protocol. Each 64-token ground-truth memorized
  sequence is split into a 32-token prompt and a 32-token target. The
  quantized model greedily generates a continuation from the prompt; an exact
  match with the target counts as successful extraction. In parallel we
  measure perplexity on held-out text at the same precision, so that lost
  memorization can be separated from lost capability.}
  \label{fig:method}
\end{figure}

Figure~\ref{fig:method} summarizes the protocol. Everything below is
deterministic given a seed, and every result file is released.

\paragraph{Models and ground truth.}
We use the deduplicated Pythia models at 160M, 410M, and 1B parameters
\citep{biderman2023pythia}. For each model we take its published set of
memorized sequences \citep{biderman2023emergent}: 64-token sequences from the
Pile that the full-precision model reproduces greedily. We sample a fixed
random subset per run and cache it, so that every precision level for a given
model is evaluated on identical sequences.

\paragraph{Extraction.}
Following the standard discoverable-extraction setup
\citep{carlini2022quantifying}, we split each 64-token sequence into a
32-token prompt and a 32-token target. We feed the prompt to the model and
greedily decode 32 new tokens with no sampling. The primary metric is the
\emph{exact-match rate}: the fraction of sequences whose generated
continuation equals the target token for token. We also record the mean
length of the correct leading prefix and the per-token accuracy, which let us
see partial forgetting rather than only the all-or-nothing outcome.

\paragraph{Capability.}
The control that makes the study meaningful is capability. At every precision
we compute perplexity with a sliding window over two corpora: WikiText-2
\citep{merity2016pointer}, which is out of the training distribution, and a
sample of Pile text, which is in it. Reporting both guards against a
capability measure that flatters or punishes quantization for
distribution-specific reasons.

\paragraph{Quantization.}
Our main precision ladder is FP32, FP16, INT8 (LLM.int8()
\citep{dettmers2022int8}), and two four-bit bitsandbytes types, NF4 and FP4
\citep{dettmers2023qlora}. To make sure any effect is a property of
quantization and not of one particular library, we add an independent method
we implement ourselves: group-wise symmetric round-to-nearest (RTN) at eight
and four bits, applied directly to the linear-layer weights, with no external
dependency. RTN is algorithmically unrelated to bitsandbytes' normal-float
scheme, so agreement between them is strong evidence that the effect is real.

\paragraph{Selectivity.}
To compare how fast memorization and capability decay, we define, relative to
a full-precision reference, the retained fractions
\[
  m(q) = \frac{\text{extraction}(q)}{\text{extraction}(\text{ref})},
  \qquad
  c(q) = \frac{\text{perplexity}(\text{ref})}{\text{perplexity}(q)},
\]
so that both are one at the reference and smaller when the model is worse,
and a \emph{selectivity}
\[
  s(q) = \frac{\log m(q)}{\log c(q)}.
\]
Selectivity greater than one means memorization is decaying faster than
capability; around one means they decay together; less than one means
capability goes first. It is simply the ratio of the two log-losses, and it
is scale-free, which is what lets us compare across precisions and models.

\section{Results}
\label{sec:results}

Table~\ref{tab:main} gives the full grid. Figure~\ref{fig:curve} shows
extraction against precision for all three model sizes, and
Figure~\ref{fig:plane} places every configuration on the
capability--memorization plane.

\begin{table}[t]
  \caption{Extraction and capability across precision and model size.
  ``Extract'' is the verbatim exact-match rate; perplexity is on WikiText-2
  (WT) and Pile; $m$, $c$, $s$ are memorization retained, capability retained
  (Pile), and selectivity relative to the full-precision reference. The
  reference is FP32 (160M, 410M) or FP16 (1B, which does not fit in FP32 on
  our GPU). RTN is our independent quantizer.}
  \label{tab:main}
  \centering
  \small
  \begin{tabular}{llccccc}
    \toprule
    Model & Precision & Extract & PPL (WT) & PPL (Pile) & $m$ & $c$ / $s$ \\
    \midrule
    \multirow{5}{*}{160M}
      & FP32 & 0.756 & 25.60 & 12.27 & 1.000 & 1.000 / --- \\
      & FP16 & 0.746 & 25.84 & 12.37 & 0.987 & 0.992 / --- \\
      & INT8 & 0.734 & 26.07 & 12.48 & 0.971 & 0.983 / 1.7 \\
      & NF4  & 0.294 & 38.39 & 17.33 & 0.389 & 0.708 / 2.7 \\
      & FP4  & 0.176 & 48.61 & 21.22 & 0.233 & 0.578 / 2.7 \\
    \midrule
    \multirow{6}{*}{410M}
      & FP32 & 0.834 & 15.86 & 8.88 & 1.000 & 1.000 / --- \\
      & FP16 & 0.822 & 15.87 & 8.89 & 0.986 & 0.999 / --- \\
      & INT8 & 0.810 & 16.13 & 8.98 & 0.971 & 0.989 / 2.6 \\
      & NF4  & 0.228 & 32.40 & 15.35 & 0.273 & 0.579 / 2.4 \\
      & RTN4 & 0.176 & 37.14 & 18.49 & 0.211 & 0.480 / 2.1 \\
      & FP4  & 0.132 & 39.60 & 20.59 & 0.158 & 0.431 / 2.2 \\
    \midrule
    \multirow{5}{*}{1B}
      & FP16 & 0.830 & 12.68 & 7.58 & 1.000 & 1.000 / --- \\
      & INT8 & 0.826 & 12.72 & 7.60 & 0.995 & 0.997 / 1.5 \\
      & NF4  & 0.596 & 13.30 & 7.90 & 0.718 & 0.959 / 8.0 \\
      & RTN4 & 0.498 & 14.28 & 8.32 & 0.600 & 0.910 / 5.4 \\
      & FP4  & 0.460 & 14.25 & 8.32 & 0.554 & 0.911 / 6.3 \\
    \bottomrule
  \end{tabular}
\end{table}

\subsection{Quantization forgets memorization faster than capability}

The clearest way to see the main effect is Figure~\ref{fig:plane}, where
every point is one model at one precision, placed by how much capability it
kept (horizontal) and how much memorization it kept (vertical). If the two
decayed together, points would sit on the diagonal. They do not. Every
quantized configuration falls \emph{below} the diagonal: it lost more
memorization than capability. The selectivity column of Table~\ref{tab:main}
says the same thing numerically, staying above one throughout.

The effect is small at eight bits, where almost nothing is lost on either
axis, and it becomes dramatic at four bits. At 410M, dropping to NF4 costs
a bit under half the capability (Pile perplexity rises by about seventy
percent) but nearly three-quarters of the memorization. The partial-match metrics tell a
consistent story: the mean correct-prefix length collapses from about 28 of
32 tokens at full precision to roughly 12 at NF4, so quantization is not just
flipping a few borderline sequences, it is shortening how far the model can
faithfully reproduce a passage before it drifts.

\subsection{But it is not a privacy defense, and gets less so with scale}

The second finding is what stops this from being a happy story for privacy.
Look at the 1B rows. Going to NF4 there costs only about four percent of
capability, yet the model still reproduces 72\% of the memorized sequences it
knew at FP16. The selective forgetting is still there, and selectivity is
actually highest at this size, around 8, but that high selectivity sits right
next to heavy leakage, because the model is barely damaged overall. The two
are not in conflict: the model loses a lot of memory relative to the tiny
amount of capability it loses, and it still had most of its memory to begin
with.

The trend across the three sizes is the uncomfortable part. At four-bit NF4,
the fraction of memorized sequences that survive quantization is 0.39 at
160M, 0.27 at 410M, and 0.72 at 1B. Read alongside the capability numbers,
the picture is that larger models absorb quantization noise so well that
their memorized content rides through almost intact. The models people
actually deploy are one to two orders of magnitude larger still. Nothing in
our data suggests the leaked fraction will conveniently fall again at that
scale; if anything it points the other way.

\subsection{The effect is not an artifact}

\begin{figure}[t]
  \centering
  \includegraphics[width=0.72\linewidth]{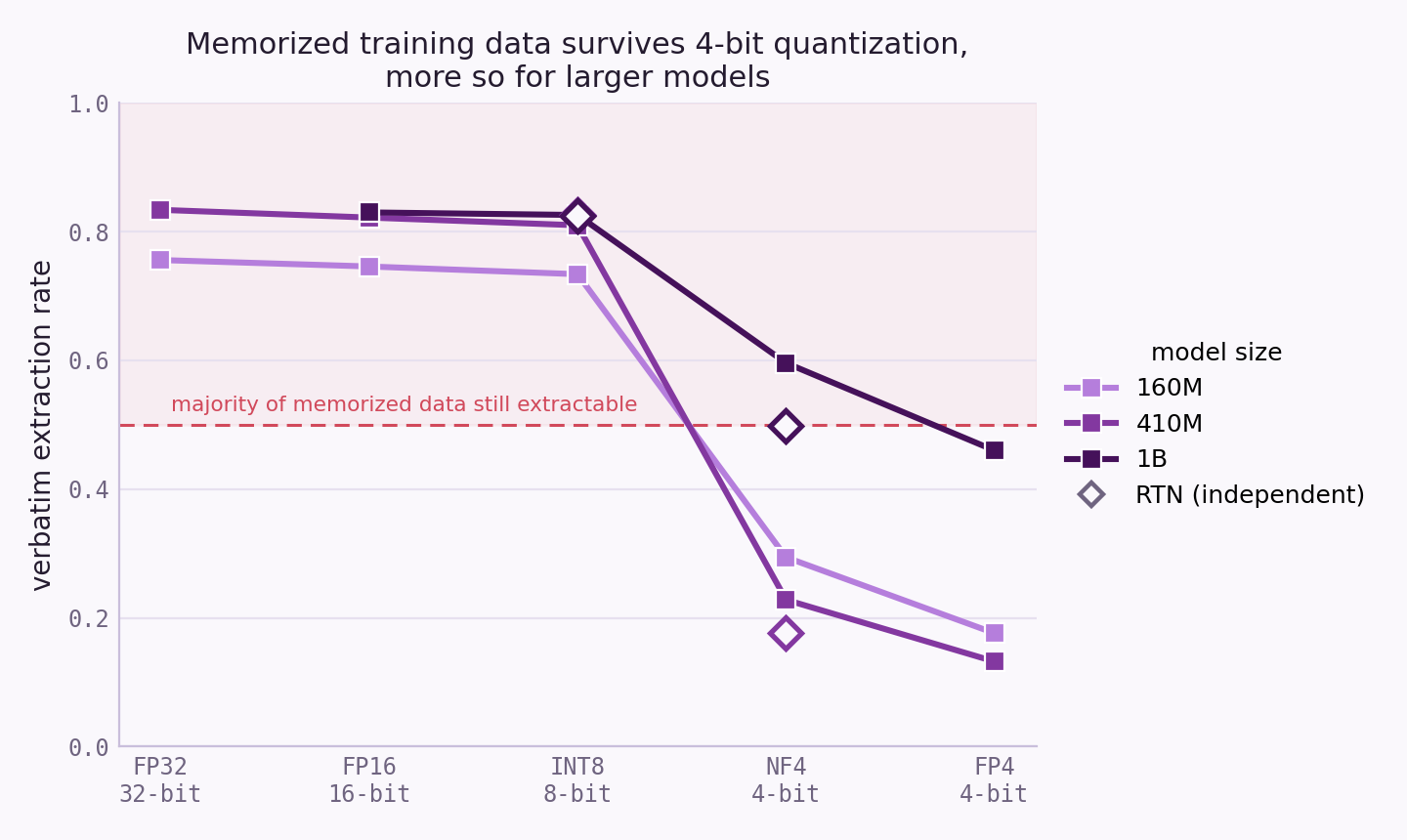}
  \caption{Verbatim extraction against precision, for the three model sizes.
  Filled squares are the bitsandbytes ladder; open diamonds are our
  independent RTN quantizer at matching bit widths. Extraction is stable down
  to eight bits and then falls at four bits, far more steeply for the smaller
  models than for the 1B model. The shaded band marks where the majority of
  memorized data is still extractable.}
  \label{fig:curve}
\end{figure}

\begin{figure}[t]
  \centering
  \includegraphics[width=0.6\linewidth]{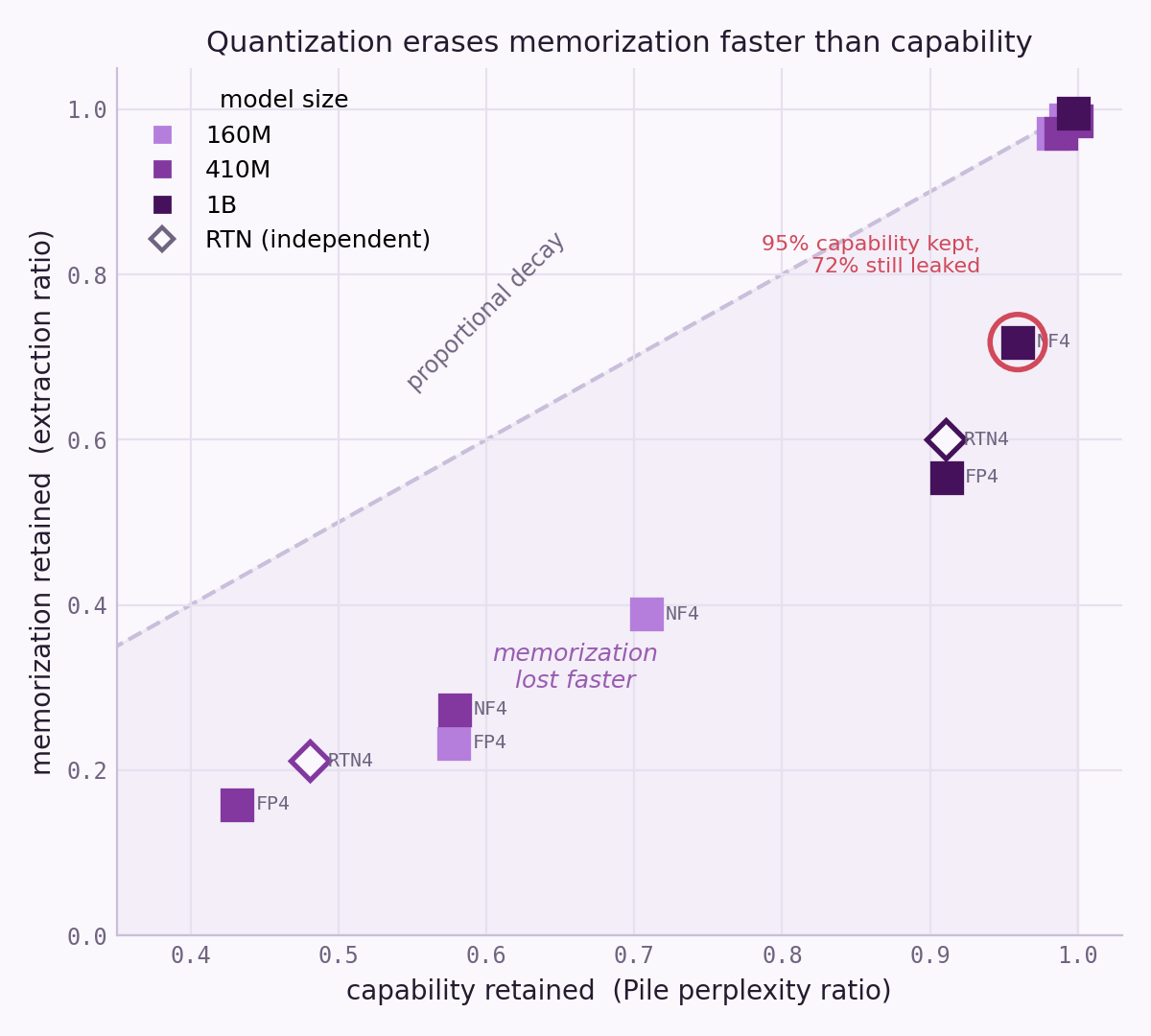}
  \caption{The capability--memorization plane. Each point is one model at one
  precision, relative to full precision. Filled squares are bitsandbytes
  quantization, open diamonds are our independent RTN method. The dashed
  diagonal is proportional decay; every quantized point lies below it, meaning
  memorization was lost faster than capability. The 1B four-bit points sit far
  to the right (little capability lost) yet well below the diagonal (much
  memorization lost), which is exactly the ``selective but not sufficient''
  regime.}
  \label{fig:plane}
\end{figure}

Two controls guard the conclusions. First, method: our hand-written RTN
quantizer, which shares no code and no algorithm with bitsandbytes,
reproduces the same collapse. At 410M, four-bit extraction is 0.228 under NF4
and 0.176 under RTN4; at 1B it is 0.596 under NF4 and 0.498 under RTN4. The
open squares in Figures~\ref{fig:curve} and~\ref{fig:plane} are the RTN
points, and they land right among the bitsandbytes points. The precise
numbers differ, as one would expect from different rounding schemes, but the
selective-forgetting pattern is identical. Second, corpus: the selectivity
computed on in-distribution Pile perplexity is if anything slightly stronger
than on WikiText, so the effect is not an artifact of measuring capability on
out-of-distribution text.

\section{Discussion and limitations}

\paragraph{What this means for practice.}
The direct consequence is a caution. If a team quantizes a model and reasons
that its memorized training data is now safe, our results say that reasoning
is wrong, and increasingly wrong the larger the model. Quantization removes
some memorized content, but at deployment scale it leaves most of it
extractable while barely denting capability. If you actually need to remove
memorized data, quantization is not the tool; targeted unlearning or a
training-time fix is, and even unlearning can be undone by a later
quantization step \citep{zhang2024catastrophic}. The second practical
point is about measurement. Because membership inference and extraction
respond differently, and because extraction is the threat with legal and
reputational teeth, we argue extraction is the metric to report when making
any privacy claim about a compressed model.

\paragraph{What this hints at scientifically.}
That memorization is more precision-fragile than capability, and that the gap
widens with scale, is a clue about where memorized content lives. A plausible
reading is that verbatim memories depend on specific, finely tuned weight
configurations that quantization noise disrupts, whereas general capability
is spread more redundantly and survives rounding. The scale trend then says
that larger models have enough redundancy to protect even those fine
configurations. This lines up with the finding that quantization can restore
unlearned knowledge \citep{zhang2024catastrophic}, and it suggests
memorization and generalization are, at least in part, physically separable
in the weights. We think this is worth following up with targeted
interpretability.

\paragraph{Limitations.}
Our results are scoped, and we state the scope plainly. We use one model
family, Pythia, because it is the only one with public training data and
public ground-truth memorization; whether the numbers transfer to other
families is open. Our largest model is 1B, well below deployment scale, so
the scale trend is an extrapolation, not a measurement at the sizes that
matter most. We study English text, weight-only post-training quantization,
and strict greedy extraction; sampling-based and probabilistic extraction
\citep{hayes2024measuring}, activation quantization, and quantization-aware
training may behave differently. Finally, extraction rate on the known
memorized set measures how quantization affects \emph{already-memorized}
content; it does not speak to what a fresh model quantized before any
memorization measurement would do. These are the natural next experiments,
and the released code is set up to run them.

\section{Conclusion}

We measured what quantization does to the training data a language model has
memorized, using real models and ground-truth memorized sequences rather than
proxies. Quantization forgets memorization faster than it forgets capability,
consistently, under two independent quantizers and two corpora. But it does
not forget enough: at the largest scale we can run, four-bit models still
reproduce most of what they memorized while losing almost no capability, and
the surviving fraction grows with model size. Compression is not a privacy
defense, and extraction, not membership inference, is the number to watch. We
release all code, sampled data, and results to make the follow-up
experiments, especially at larger scale, easy to run.

\bibliographystyle{plainnat}

\appendix

\section{Reproducibility}

All numbers come from seeded runs over the public Pythia memorized-evals set.
The sampled evaluation sequences are cached and released so that every
precision level is scored on identical data. The independent RTN quantizer is
a few lines of group-wise round-to-nearest applied to the linear weights and
has no external dependency. Code is at
\url{https://github.com/AkshaySasi/bits-and-memories}, and the sampled
sequences and per-configuration result files are at
\url{https://huggingface.co/datasets/AkshaySasi/bits-and-memories}.

\section*{Acknowledgements}

AI assistance was used in developing the experiment code and in drafting and
editing the manuscript. All experiments were designed and executed by the
author, and all results, claims, and references were verified by the author.

\end{document}